\documentclass[twoside]{article}

\usepackage{PRIMEarxiv}

\usepackage[utf8]{inputenc} % allow utf-8 input
\usepackage[french]{babel}
\usepackage[T1]{fontenc}    % use 8-bit T1 fonts

\usepackage{hyperref}       % hyperlinks
\usepackage{url}            % simple URL typesetting
\usepackage{booktabs}       % professional-quality tables
\usepackage{amsfonts}       % blackboard math symbols
\usepackage{nicefrac}       % compact symbols for 1/2, etc.
\usepackage{microtype}      % microtypography
\usepackage{fancyhdr}       % header
\usepackage{graphicx}       % graphics
 \usepackage{subcaption}
 \usepackage{xcolor}
 \usepackage{multirow}
 \usepackage{amsmath}
 \usepackage{tabulary}
 \newcolumntype{K}[1]{>{\centering\arraybackslash}p{#1}}
 
\graphicspath{{./}}     % organize your images and other figures under media/ folder

\title{{\it GenderedNews}: Une approche computationnelle des écarts de représentation des genres dans la presse française
}

\author{
Richard, A. \\
  Univ. Grenoble Alpes, \\
  PACTE, LIG \\
  38000 Grenoble, France\\
  \texttt{ange.richard@univ-grenoble-alpes.fr} \\
  \And
  Bastin, G. \\
  Univ. Grenoble Alpes, CNRS,\\
  Sciences Po Grenoble, PACTE,\\
  38000 Grenoble, France \\
  \texttt{gilles.bastin@iepg.fr} \\
  \And
  Portet, F. \\
  Univ. Grenoble Alpes, CNRS,\\
  Grenoble INP, LIG\\
   38000 Grenoble, France \\
   \texttt{francois.portet@imag.fr} \\ 
}

\begin{document}
\maketitle

\begin{abtract}
In this article, we present {\it GenderedNews} (\url{https://gendered-news.imag.fr}), an online dashboard which gives weekly measures of gender imbalance in French online press. We use Natural Language Processing (NLP) methods to quantify gender inequalities in the media, in the wake of global projects like the Global Media Monitoring Project. Such projects are instrumental in highlighting gender imbalance in the media and its very slow evolution. However, their generalisation is limited by their sampling and cost in terms of time, data and staff. Automation allows us to offer complementary measures to quantify inequalities in gender representation. We understand representation as the presence and distribution of men and women mentioned and quoted in the news -- as opposed to representation as stereotypification. In this paper, we first review different means adopted by previous studies on gender inequality in the media : qualitative content analysis, quantitative content analysis and computational methods. We then detail the methods adopted by {\it GenderedNews} and the two metrics implemented: the masculinity rate of mentions and the proportion of men quoted in online news. We describe the data collected daily (seven main titles of French online news media) and the methodology behind our metrics, as well as a few visualisations. We finally propose to illustrate possible analysis of our data by conducting an in-depth observation of a sample of two months of our database.
\end{abtract}
~\\

\begin{resume}
Nous présentons dans cet article {\it GenderedNews} (\url{https://gendered-news.imag.fr}), un tableau de bord en ligne permettant de mesurer les écarts de représentation des genres dans la presse française en ligne. Nous utilisons des méthodes automatiques du traitement automatique des langues (TAL) pour quantifier les inégalités de genre dans les médias, en complémentarité des projets de grande envergure tels que le Global Media Monitoring Projet. Ces études ont mis en lumière le déséquilibre de genres dans les médias et l'inertie de son évolution. Cependant, leur échantillonnage et leur coût en termes de temps, de données et de main-d'œuvre restent des limites à leur généralisation. L'approche automatique nous permet de proposer des mesures complémentaires pour quantifier les inégalités de genre dans les représentations médiatiques. Nous entendons la notion de représentation comme la présence et la distribution des hommes et des femmes mentionnés $\cdot$es et cité$\cdot$es dans les médias -- et non la représentation au sens de stéréotypisation. Dans cet article, nous commençons par faire une revue des différentes manières utilisées dans la littérature étudiant les écarts de genre dans les médias: l'analyse de contenu qualitative, l'analyse de contenu quantitative, et l'analyse computationnelle de contenu ({\it text mining}). Nous détaillons ensuite les méthodes adoptées pour {\it GenderedNews} et les deux mesures que nous avons implémentées: le taux de masculinité des mentions et la proportion des hommes cités. Nous décrivons les données que nous collectons tous les jours (les articles de sept des titres principaux de la presse d'information en ligne française), la méthodologie de nos mesures, ainsi que quelques visualisations du tableau de bord. Nous terminons en proposant une observation plus poussée d'un échantillon de deux mois de notre base de données.
\end{resume}

% keywords can be removed
\renewcommand{\keywordname}{{\bfseries \emph Mots-clés}}

%Introduction
\section{Introduction}
\label{sec:intro}
    
    \paragraph{}{Dans son rapport récent sur la place des femmes dans les médias en temps de crise, Céline Calvez constate que la crise sanitaire de 2020 a causé un recul net de la présence des femmes dans les médias. Celles qui étaient présentes dans les médias étaient représentées plus souvent comme des témoins passifs, souvent anonymes, que comme des actrices ou des expertes de la pandémie. Pourtant, comme le souligne aussi ce rapport, les femmes constituent une des catégories qui ont été le plus touchées par la pandémie, tant sur le plan professionnel (métiers du {\it care}, métiers dits ‘essentiels'\dots) que social (les inégalités au sein du foyer, avec le travail domestique et la garde d'enfants, ainsi que les violences domestiques, ont augmenté durant les confinements successifs). Les personnes les plus vulnérables à la crise ont donc été, en grande partie, effacées du débat public et de la scène médiatique \cite{calvez_remise_2020}.}
    
    \paragraph{}{Cette crise a exacerbé le phénomène d'inégalités de genre dans les médias. Ce phénomène est connu et observé de longue date dans les sciences sociales sous deux formes: celle des écarts de représentation (les femmes sont moins souvent représentées dans les médias) et celle des biais ou stéréotypes de représentation (lorsqu'elles sont représentées, leur personnalité ou leurs actions sont souvent stéréotypées). Durant ces deux dernières décennies, un nombre grandissant d'études visant à démontrer les écarts et biais de représentation en fonction du genre dans les médias d'information ont été publiées. Ces études confirment, malgré des progrès lents constatés de la part des femmes mentionnées ou citées, le diagnostic d' "annihilation symbolique" formulé dans les années 1970 \cite{tuchman_symbolic_1978}. Pour ne citer qu'un exemple, le {\it Global Media Monitoring Project}, qui mesure tous les cinq ans depuis 1995 la présence des femmes dans les médias à partir d'une analyse de contenu menée le même jour dans un grand nombre de pays, rapporte une part de présence de seulement 29 \% des sujets et sources mentionnés en 2020 \cite{macharia_who_2021}, soit une augmentation de seulement 13 points en vingt-cinq ans. Ces études ont permis de mettre en lumière l'inertie de l'évolution de ces écarts. Elles ont aussi permis de porter le déséquilibre de représentation au devant du débat public. Ces études sont cependant limitées par leur échantillonage (souvent fait sur une journée ou une courte période) et par le coût de leur codage manuel.}
    
    \paragraph{}{Le projet {\it GenderedNews} que nous présentons dans cet article a pour vocation de proposer une manière complémentaire de quantifier les écarts de genre dans les médias par rapport aux enquêtes de grande envergure fondées sur l'analyse de contenu. Notre but est de fournir un outil permettant la mesure des écarts de représentation dans la presse française de manière automatique et continue sur un grand nombre de sources. Cette démarche a pour intérêt notamment de neutraliser les effets de contexte liés à des corpus collectés de manière ponctuelle. Elle repose sur l'utilisation des méthodes de traitement automatique de la langue (TAL) afin de calculer dans chaque article du corpus étudié la part des hommes et des femmes mentionnées (dont le nom apparaît) et celle des hommes et des femmes citées (à qui la parole est donnée). Les résultats de ces mesures, dont nous détaillons la construction dans la suite de cet article, sont disponibles sur le site {\it web} correspondant à ce projet \footnote{\url{https://gendered-news.imag.fr/}} qui est actualisé toutes les semaines. Le but du site {\it GenderedNews} est de fournir des données et des mesures originales et régulières pour mieux comprendre le phénomène d'inégalité genrée des contenus journalistiques, et son évolution dans le temps et entre différents médias français.}

%%%%%%%% Revue de Littérature
\section{Travaux connexes}
\label{sec:litt}

        \paragraph{}{De nombreux travaux existent sur l'étude des inégalités de représentation dans les médias selon un large panel de critères et d'appartenances sociales. La question du genre est cependant celle qui a reçu le plus d'attention dans la littérature.}
        
        \paragraph{}{La recherche sur les biais de genre dans les médias a principalement emprunté deux voies complémentaires : la mesure de l'écart quantitatif de représentation entre femmes et hommes ({\it "gender imbalance"}) et l'analyse qualitative des stéréotypes de genre ({\it "gender stereotyping"}). Dans le contexte des années 1970 marqué par des revendications croissantes d'égalité entre les hommes et les femmes et par le développement de nouveaux médias tels que les chaînes de télévision commerciale et les magazines, ces deux voies ont été explorées pour comprendre la capacité des médias à refléter, et donc à renforcer, les valeurs dominantes de la société, notamment celles relatives aux rôles sexuels des hommes et des femmes. Le déséquilibre numérique entre les hommes et les femmes dans les médias et la stéréotypisation des rôles offerts aux un$\cdot$es et aux autres sont alors analysés conjointement, comme par exemple dans l'étude des fictions télévisuelles dans lesquelles "la représentation au sein du monde de la fiction signifie l'existence; l'absence signifie l'annihilation symbolique"\footnote{Traduction personnelle. Citation d'origine: {\it "representation in the fictional world signifies social existence; absence means symbolic annihilation"}} \cite{gerbner1972symbolic}. Gaye Tuchman a systématisé cette perspective de l'"annihilation symbolique — en l'occurrence des femmes \cite{tuchman_symbolic_1978}}.
        
        \paragraph{}{Dans ce qui suit, après avoir précisé le sens que nous donnons au concept de genre, nous rappelons les principales méthodes qui ont été utilisées pour mesurer l'écart de représentation entre hommes et femmes dans les médias, à savoir l'analyse de contenu qualitative, l'analyse de contenu quantitative, et l'approche computationnelle. Cette revue de littérature nous permet de montrer l'originalité de l'approche retenue pour le projet {\it GenderedNews} qui relève de l'analyse computationnelle.}

    \subsection{Précisions préliminaires sur le concept de genre}
    \label{sec:defgenre}
    
         \paragraph{}{Il nous est d'abord nécessaire de définir la manière dont nous considérons la notion de genre au sein de cette étude. Le genre, ici, est compris comme une construction sociale, culturelle et historique, qui divise socialement les individus selon une catégorisation binaire dans les sociétés occidentales. West et Zimmerman définissent le genre comme "un trait émergent des situations sociales: à la fois le résultat de et une logique pour des arrangements sociaux divers, et le moyen de légitimer une des divisions les plus fondamentales de la société" \footnote{Traduction personnelle. Citation d’origine: {\it "an emergent feature of social situations: both as an outcome of and a rationale for various social arrangements and as a means of legitimating one of the most fundamental divisions of society"}} \cite[p.126]{west_doing_1987}. La construction de cette catégorisation se fonde sur une différence perçue entre les sexes \cite{scott_gender_1986} qui est naturalisée dans la société, c'est-à-dire que son caractère construit est invisibilisé \cite{beauv_1949} dans toute une série d'artefacts culturels fondés sur la reproduction de l'opposition binaire et hiérarchisée entre hommes dominants et femmes dominées (comme la littérature, le cinéma, la publicité et l'information). Nous nous intéressons donc ici tout particulièrement aux médias, artefacts culturels qualifiés de "technologies du genre" par \cite{lauretis_technologies_1987}. L'espace médiatique est un espace fondé sur les hiérarchies de dominations genrées, au sein duquel ces catégories de genre sont construites et renforcées.}
     
    \subsection{Analyse de contenu qualitative}
    \label{sec:acqual}
   
        \paragraph{}{L'analyse de contenu qualitative est la méthode la plus fréquemment utilisée pour la mise en évidence des stéréotypes masculins et féminins dans les médias \cite{garcin2007violences, coulomb2009corps, coulomb2017femmes, matonti2017genre}. Celle-ci repose sur le dépouillement manuel de corpus de taille limitée et l'identification dans ces corpus — à l'aide de méthodes favorisant l'induction ou la déduction de catégories \cite{altheide1996process, mayring_qualitative_2000} — de stéréotypes récurrents appliqués au "déploiement" ({\it display}) et à la "stylisation" ({\it styling}) des identités de genre \cite{goffman_gender_1979}. Cette méthode, souvent, ne permet qu'un contrôle limité de l'échantillonnage du corpus étudié, un choix restreint des catégories et une significativité limitée des résultats obtenus. Cependant, elle est adaptée à la recherche de stéréotypes latents qui ne sont pas, ou pas facilement, réductibles à des formes langagières univoques ou n'étant pas susceptibles d'être comptées. Elle permet également de mener des études de cas dans lesquelles la question de la comparaison des résultats est peu pertinente, ou portant sur des corpus complexes contenant par exemple à la fois des textes et des images. Elle présente aussi l'intérêt d'être facilement accessible, raison pour laquelle elle a été utilisée au-delà de la recherche universitaire \cite{tuchman_womens_1979}.}
        
        \paragraph{}{Ces méthodes ont notamment permis de mettre en évidence le fait que dans les médias, les femmes sont plus souvent caractérisées par des attributs physiques (mention du corps, du vêtement, etc.) que les hommes, qu'elles sont plus souvent caractérisées par leur statut matrimonial (femme de ou mère/fille de), qu'elles sont plus souvent présentées comme victimes et non comme actrices des problèmes sociaux.}
    
    \subsection{Analyse de contenu quantitative}
    \label{sec:acquant}
    
       \paragraph{}{L'analyse de contenu quantitative a été utilisée pour examiner l'autre versant de l'idée d'annihilation symbolique, à savoir l'écart de représentation, qui nous intéresse principalement ici. De nombreuses recherches ont utilisé cette méthode depuis les années 1970 pour mettre en évidence le niveau élevé des écarts de représentation entre hommes et femmes dans les médias. Si la taille des corpus étudiés ne varie pas sensiblement par rapport à l'approche précédente, il s'agit alors de faire des mesures systématiques de la place occupée par les hommes et les femmes dans les contenus. Plusieurs stratégies empiriques ont été mises en œuvre dans ce but, que ce soit en termes de choix de corpus de référence ou en termes de manières de mesurer la place des hommes et des femmes. La première approche est la plus proche de la proposition de Tuchman selon laquelle l'écart de représentation trouve son origine dans la façon dont l'ensemble de la société est "reflétée" dans les médias \cite[p.7]{tuchman_symbolic_1978}. Elle consiste à étudier des corpus de contenus médiatiques abordant tous les aspects de la vie de la société sur une période donnée \cite{kinnick_gender_1998, mace2006societe, ross_gender_2012, coulomb2018affaire, ojebuyi_gender_2018, mitchelstein_joanne_2019}. Plusieurs études américaines à la fin des années 1970 et au début des années 1980 ont par exemple montré que les femmes représentent moins de 10\% des personnes citées dans les articles des principaux journaux \cite{davis_sexist_1982, potter_gender_1985}. Une étude basée sur le contenu du New York Times en 1885 et 1985 a démontré quant à elle que malgré une réduction des stéréotypes de genre dans le journal, le déséquilibre entre les genres n'a pas changé de façon spectaculaire au cours de la période étudiée, pourtant très longue \cite{jolliffe_comparing_1989}. Une autre étude portant sur 18 journaux nationaux et locaux américains en 1999 a conduit à estimer que les femmes sont en moyenne représentées trois à quatre fois moins que les hommes dans ces médias \cite{armstrong_influence_2004}.}
       
       \paragraph{}{Cette méthode est privilégiée dans de nombreux rapports sur la place des femmes dans les médias par exemple : dans le rapport précurseur de la US Commission on Civil Rights \footnote{\url{https://www2.law.umaryland.edu/marshall/usccr/documents/cr12t23.pdf}}, le rapport de Michèle Reiser et Brigitte Grésy sur l'image des femmes dans les médias français \cite{reiser_rapport_2008, reiser_les_2011}, ou plus récemment celui de Céline Calvez sur la place des femmes dans les médias en temps de crise \cite{calvez_remise_2020}, ou encore dans le baromètre de la diversité que le Conseil Supérieur de l'Audiovisuel publie tous les ans depuis 2009.}
       
       \paragraph{}{Le Global Media Monitoring Project (GMMP) est aujourd'hui le projet le plus ambitieux visant à mesurer l'écart moyen de représentation à l'échelle internationale avec des méthodes relevant de l'analyse de contenu quantitative. Il est fondé sur l'analyse du contenu d'une journée d'activité médiatique dans le monde tous les cinq ans depuis 1995 (\cite{gmmp_global_1995, george_who_2000, gallagher_who_2005, macharia_who_2015}). Ses résultats montrent qu'après une phase de réduction de l'écart de représentation comme des stéréotypes dans les années 1990 et le début des années 2000, la période qui suit est marquée par une stabilisation étonnante qui a été qualifié de "glaciale" (\cite{macharia_who_2021}), voire dans certains cas par une légère baisse de la représentation des femmes dans les médias. La proportion de femmes dans les médias dépouillés à l'échelle internationale est passée de 17\% en 1995 à 24\% en 2010 mais qu'elle n'a pas augmenté sensiblement depuis cette date jusqu'en 2020 où a été noté un "progrès sans révolution" de 1\% de pourcentage \cite{macharia_who_2021}.}
       
       \paragraph{}{L'intérêt d'une approche quantifiée des contenus en matière de mesure de l'écart de représentation tient dans la possibilité de comparaisons internationales (pas de dépendance à la langue), la possibilité de comparaisons dans le temps, la production de résultats agrégés simples à comprendre et efficaces politiquement (par exemple le pourcentage de femmes dans les médias). La méthode a en revanche des inconvénients: la dépendance à la période très courte retenue pour le codage (les effets de contexte massifs pour le GMMP sont discutés par Coulomb-Gully et Meadel \cite{meadel_plombieres_2011}), la dépendance aux sources retenues (elles sont souvent limitées aux médias considérés comme les plus "importants" et, en ce qui concerne la presse, l'étude ne porte que sur les douze premières pages des journaux), la qualité variable du codage manuel (en général sous-traité ou réalisé par des équipes hétérogènes: militant$\cdot$es, journalistes, universitaires). Une autre limite concerne l'approche d'agrégation au niveau national \footnote{Sur les limites du nationalisme méthodologique, voir \cite{wimmer2002methodological, beck2007cosmopolitan}}.}
       
       \paragraph{}{Pour pallier ces limites, certains domaines particuliers de la vie sociale ont été plus étudiés que d'autres, comme le sport ou la politique, dans la mesure où l'exposition médiatique y joue un rôle particulièrement important et détermine en partie les ressources et les opportunités des individus en son sein. Se concentrer sur des groupes restreints d'individus permet aussi d'échapper à une critique possible des approches globales, à savoir le manque de contrôle sur les variables socio-démographiques susceptibles d'expliquer la sur-exposition des hommes dans les médias. En mesurant l'ensemble des femmes et des hommes mentionné$\cdot$es ou cité$\cdot$es dans un journal, le risque est en effet de mesurer, au-delà de l'appartenance de genre, des phénomènes liés à l'ancienneté dans un domaine, au pouvoir ou à la notoriété. Les études les plus récentes qui ont adopté ce principe méthodologique pour l'étude de la médiatisation des hommes et des femmes dans des sphères d'activité restreintes comme la politique ont permis de mettre en évidence l'existence d'un phénomène de sous-représentation nette des femmes. Dans une étude sur la médiatisation des membres du parlement belge à la télévision en langue flamande entre 2003 et 2011, il a été démontré que l'écart de représentation entre hommes et femmes subsiste même si l'on contrôle l'âge et la position au parlement, dans les partis politiques ou au gouvernement \cite{hooghe_enduring_2015}. Une autre étude prenant en compte 20000 individus dans un grand nombre de professions aux États-Unis a permis de confirmer l'existence d'un biais net de représentation une fois contrôlés l'âge, la position sociale et la visibilité publique sur Wikipedia \cite{shor_large-scale_2019}.}

    \subsection{Approche computationnelle}
    \label{sec:comput}

        \paragraph{}{Le développement du TAL et la mise à disposition de bases de données de plus en plus volumineuses de contenus médiatiques ont conduit depuis quelques années au développement d'une approche "computationnelle" d'analyse de contenu. %C'est une approche que nous qualifions de "computationnelle" de la mesure des écarts de genre. 
        Cette approche se distingue des précédentes par l'adoption de technologies automatiques permettant d'effectuer des études sur des très grands volumes de données.}
        
        \paragraph{}{Pour la presse écrite, la grande majorité des études portent sur des journaux anglophones. Les mesures sont souvent fondées sur un lexique préétabli ou sur la reconnaissance d'entités nommées ({\it Named Entity Recognition}), qui consiste à extraire automatiquement les expressions se référent à des entités individuelles comme les personnes, fictives ou réelles, ou les noms de lieux) et l'attribution d'un genre à celles-ci \cite{jia_women_2016,shor_large-scale_2019,Dacon2021}. \cite{jia_women_2016} observent par exemple sur un corpus de plus de 2,3 millions d'articles de presse anglaise en ligne qu'un nom de personne mentionné dans un article compte entre 69,5\% (Divertissement) et 91,5\% (Sports) de chances d'appartenir à un homme selon les sujets, et que les images ont entre 59,3\% (Divertissement) et 79,9\% (Politique) de chances de représenter un homme. \cite{garg_word_2018} proposent de quantifier grâce à la technique du ‘plongement de mots' ({\it word embeddings}, une méthode de représentation lexicale par vectorisation) l'évolution des biais genrés et des stéréotypes raciaux sur une période d'un siècle dans les journaux américains. Une autre étude \cite{shor_paper_2015} sur les principaux journaux américains et un grand nombre de sources numériques a par exemple permis de mesurer l'évolution de la place des femmes dans ces médias à partir des noms de personnes mentionnées entre 1983 et 2008. La part des femmes a évolué positivement dans ces médias mais à un rythme faible (elle est passée de moins de 20 pourcent à 27 pourcent). Les auteurs de l'étude mettent ainsi en évidence l'existence d'un {\it "paper ceiling"} en matière de féminisation des contenus médiatiques. Celui-ci peut être expliqué selon eux par une variable macro-sociologique, à savoir la stagnation de nombreux indicateurs sur la place des femmes dans la société pendant cette période\footnote{Plusieurs indicateurs d'égalité genrée sur de nombreux plans de la vie sociale ont cessé de progresser sensiblement dans les années 2010, sauf dans le domaine de l'accès aux positions de pouvoir politique et économique (voir \url{https://eige.europa.eu/publications/gender-equality-index-2020-key-findings-eu.})}. Les technologies du TAL permettent également de mener des études sur les sources audiovisuelles grâce aux systèmes de reconnaissance du genre des locutrices et locuteurs. En France, par exemple, l'Institut National de l'Audiovisuel a mené une étude d'envergure sur la radio et la télévision à l'aide d'outils de reconnaissance de la parole \cite{doukhan_open-source_2018}. Pour les états-unis, des recherches ont utilisé des techniques d'analyse d'image et de TAL pour analyser 20 ans de publications des sites du  New York Times et de Fox News \cite{ash2021visual}.}
        
        \paragraph{}{Ces méthodes cependant présentent des limites et inconvénients par rapport à l'analyse de contenu quantitative évoquée dans la section \ref{sec:acquant}. Elles sont notamment très dépendantes à la langue, ce qui rend la comparaison internationale difficile. Elles prennent également appui sur des sources de données disponibles pour la recherche, qui sont limitées. La complexité des mesures produites peut également être un frein à leur diffusion. Cependant, elles présentent de nombreux avantages, en offrant la possibilité de comparer des résultats sur une très longue durée lorsque des données sont disponibles. L’automatisation des mesures permet un gain de temps et une garantie d'uniformisation liés à la suppression de la phase de codage manuel. Elle rend enfin possible et ce de manière quasi-instantanée l'analyse de très gros corpus intégrant un grand nombre de sources différentes. C'est d'ailleurs la raison de l'adoption de cette approche dans plusieurs projets d'outils automatiques mesurant la répartition genrée des sources citées dans les journaux, notamment le {\it Gender Gap Tracker} \footnote{\url{https://gendergaptracker.informedopinions.org/}} au Canada \cite{asr_gender_2021}. Pour le français, le projet Mediabot suisse \footnote{\url{https://mediabot.ch/gender-tracker}} calcule chaque semaine pour un nombre de journaux suisses en ligne la répartition du genre des personnes connues de {\it Wikipedia} dont les noms sont détectés dans les articles. Un outil similaire a été également développé pour la presse suédoise, britannique et américaine \footnote{\url{http://www.prognosis.se/GE/}, \url{http://www.prognosis.se/GE/UK/}, \url{http://www.prognosis.se/GE/USA/}}. C'est dans la lignée de ces derniers exemples que le projet {\it GenderedNews} s'inscrit.}

%%%%%%%% Approche générale
\section{Approche générale}
\label{sec:approche}

    \paragraph{}{Dans cette section, nous explicitons et justifions l'approche générale choisie dans le projet {\it GenderedNews} pour ce qui concerne le choix des médias étudiés et la définition des deux indicateurs implémentés pour l'instant permettant la mesure des écarts de représentation des genres.}
    
    \subsection{Choix des médias étudiés}
    \label{sec:titres}
        
        \paragraph{}{Les médias étudiés dans le projet {\it GenderedNews} sont les sept premiers titres de la presse quotidienne nationale payante française en termes de diffusion selon le classement de l'Alliance pour les chiffres de la presse et des médias (ACPM) sur la période 2020-2021\footnote{\url{https://www.acpm.fr/Les-chiffres/Diffusion-Presse/Presse-Payante/Presse-Quotidienne-Nationale}}. Dans l'ordre décroissant de diffusion il s'agit du Monde (424 085 exemplaires), du Figaro (338 978), de L'Équipe (209 231), des Échos (135 089), de La Croix (85 014), de Libération (83 808) et de Aujourd'hui en France (78 034), l'édition nationale du Parisien. Ces journaux ont été choisis parce qu'ils assurent une production importante et régulière d'articles d'actualité sur un grand nombre de sujets (à l'exception de L'Équipe qui ne traite que de sport, et des Echos qui traite principalement d'économie). Ils présentent aussi l'avantage d'avoir une activité notable sur les réseaux sociaux, ce qui leur assure une forte visibilité, ainsi que d'être fréquemment cités et repris par d'autres médias.}
    
        \paragraph{}{Les contenus publiés collectés proviennent des sites web de ces journaux\footnote{Il s'agit des site web suivants : \url{www.lemonde.fr}, \url{www.lefigaro.fr}, \url{www.lequipe.fr}, \url{www.lesechos.fr}, \url{www.la-croix.com}, \url{www.liberation.fr}, \url{www.leparisien.fr} (Aujourd'hui en France, édition nationale du Parisien, n'a pas de site web propre).}. Nous prenons donc en compte le contenu textuel librement accessible (sans abonnement) dans le code {\it html} des articles publiés. Celui-ci peut différer du contenu accessible dans la version papier des journaux dans la mesure où certains articles disponibles en ligne ne sont pas imprimés. La taille moyenne des textes analysés varie selon les médias étudiés. Mis à part le cas de lequipe.fr qui limite fortement cette taille, elle est comprise entre 300 et 700 mots par article (voir le tableau \ref{tab:infos_db}). Nous considérons ces portions significatives de textes non pas comme des échantillons représentatifs du reste de l'article mais comme un produit médiatique à part entière qui est le fruit de choix éditoriaux. Le texte que nous analysons est, d'une part, le début de l'article dans lequel les journalistes mentionnent les informations les plus importantes selon le principe de l'écriture en pyramide inversée. Il est aussi celui qui est rendu disponible au plus grand nombre parmi les visiteuses et visiteurs du site, puisque son accès ne nécessite aucune démarche d'abonnement préalable.}
        
        \begin{table}[h]
              \centering
              \begin{tabular}{|K{2cm}|K{2cm}|K{3cm}|K{3cm}|K{4cm}|}
                \hline
                Source  & Nombre d'articles collectés & Nombre moyen de mots par article & Taux de masculinité des mentions moyen & Proportion moyenne des hommes dans les personnes citées \\
                \hline
                la-croix.com    &   10 457  &   661.92  &   0.75    &   0.73 \\
                lefigaro.fr &   23 522  &   417.77  &   0.77    &   0.77 \\
                lemonde.fr  &   20 025  &   596.55  &   0.76    &   0.76 \\
                leparisien.fr   &   25 686  &   318.54  &   0.76    &   0.76 \\
                lequipe.fr   &  18 152  &   81.13   &   0.87    &   0.91 \\
                lesechos.fr &   5214  &   395.13  &   0.75    &   0.8 \\
                liberation.fr   &   9 941   &   541.95  &   0.73    &   0.75 \\
                \hline
              \end{tabular}
              \caption{Informations sur les articles collectés\\entre le 08/03/21 et le 22/12/21}
            \label{tab:infos_db}
            \end{table}

    \subsection{Mesures de l'écart de représentation}
    \label{sec:descrmes}
        
        \paragraph{}{Deux indicateurs ont été élaborés dans le cadre de {\it GenderedNews}: la part des hommes mentionnée par rapport à l'ensemble des femmes et hommes mentionné$\cdot$es dans l'article, que nous appelons le "taux de masculinité des mentions", et la part des hommes cités par rapport à l'ensemble des personnes cité$\cdot$es (c'est-à-dire celles et ceux à qui la parole est donnée) dans l'article. Nous calculons ces deux indicateurs de l'écart de représentation entre hommes et femmes dans les articles chaque jour pour l'ensemble des articles publiés par chacun des journaux de notre corpus. La partie qui suit décrit ce que mesure chaque indicateur. La méthodologie et la mise en oeuvre technique de ces indicateurs est explicitée plus bas dans la section \ref{sec:methodo}.}

        %% Mentions
        \subsubsection{Taux de masculinité des mentions}
            
            \paragraph{}{Une des difficultés récurrentes des recherches sur l'écart de représentation genrée dans les médias est le choix d'une méthode de codage des données. Le choix qui a été fait dans {\it GenderedNews} repose sur une méthode computationnelle visant à mesurer dans un premier temps le niveau de masculinité des mentions de chaque article \footnote{Nous entendons "masculinité" comme désignant la production dans les articles du corpus d'une sur-exposition systématique des hommes par rapport aux femmes. Nous le distinguons d'une autre utilisation de ce terme dans la littérature sur le genre qui l'associe davantage à la performance de rôles sexuels dans les interactions de la vie quotidienne \cite{connell2005masculinities}}. Il repose sur l'attribution d'un score de masculinité à chaque prénom non ambigu identifié dans un article (c'est-à-dire chaque prénom qui ne peut pas désigner autre chose qu'une personne). Ce score est égal à la probabilité que ce prénom ait été attribué à un garçon en France entre 1900 et 2017. L'indice de masculinité globale de chaque article est donc égal à la moyenne des scores de chacun des prénoms non ambigus identifiés dans l'article. Il varie de 0 (tous les prénoms identifiés sont exclusivement donnés à des filles à l'état-civil) à 1 (tous les prénoms identifiés sont exclusivement donnés à des garçons).}
            
            \paragraph{}{Cette méthode de détermination de la masculinité globale des mentions dans les articles présente plusieurs avantages par rapport au codage manuel du genre utilisé dans les études reposant sur une analyse de contenu des articles. Le plus évident est la possibilité de l'appliquer assez simplement à de vastes corpus de textes tels que celui utilisé dans cet article. Du fait du nombre important d'articles collectés dans notre projet, il est en effet difficile d'envisager un codage manuel systématique du genre des personnes mentionnées. La méthode retenue s'appuie sur des outils du traitement automatique des langues naturelles qui permettent de rendre cette opération transparente et mécanique. Un second avantage consiste dans le fait que cette méthode n'assigne pas un genre binaire aux différents prénoms rencontrés dans les articles mais un score continu de masculinité dont la valeur repose sur le niveau d'association au masculin des prénoms attribués aux enfants nés en France. Certaines limites du codage manuel du genre des personnes mentionnées dans un article peuvent ainsi être dépassées comme par exemple le codage des prénoms épicènes. Dans ce cas particulier, la méthode, si elle s'appuie sur la régularité statistique pour genrer les prénoms épicènes plus que sur la réalité empirique, permet de se prémunir du risque d'assignation de genre lié au codage humain dans la mesure où les codeurs ou codeuses sont susceptibles de s'appuyer sur des préjugés ou sur des compétences inégalement réparties à identifier le genre d'une personne en fonction d'indices contenus dans un texte.}
            
           Le calcul du taux de masculinité pour un article est défini par l'équation \ref{eq:taux_masc}. Pour un document (article) $d$, le taux $Taux\_masc_d$ est calculé par la somme des $m_i$ divisée par le nombre total $N$ de prénoms trouvés dans le document $d$. $m_i \in [0,1]$ représente le score de masculinité du $i^e$ prénom. Un $m_i=0$ signifie que le $i^e$ prénom n’est attribué qu’à des femmes tandis que $m_i=0.5$ est caractéristique d'un prénom épicène (attribué autant à des femmes qu'à des hommes). Le taux de masculinité pour une source donnée de documents $D$ est calculée par l'équation \ref{eq:taux_masc_source}.  Dans notre cas, nous nous intéressons à une mesure quotidienne de chaque site web de journal, $D$ est donc constitué de tous les articles de journaux du jour.  
           
           \begin{equation}\label{eq:taux_masc}
            Taux\_masc_d =\frac{1}{N} \sum_i^{N} m_i , \quad \text{avec} \quad m_i \in [0,1]
            \end{equation}
            
            \begin{equation}\label{eq:taux_masc_source}
            Taux\_masc(D) =\left( \frac{1}{|D|} \sum_d Taux\_masc_d \right) \in [0,1]
            \end{equation}

            \paragraph{}{Cette méthode présente cependant aussi des inconvénients comme le fait qu'elle conduit à coder comme des prénoms certains mots qui n'en sont pas malgré le travail de désambiguation mené sur la base de données utilisée pour extraire les prénoms. C'est notamment le cas des noms de famille qui sont aussi des prénoms comme Martin. Le défaut le plus important de la méthode est cependant qu'elle ne permet pas de distinguer les personnes mentionnées et donc de collecter des informations à leur sujet permettant de les différencier dans l'analyse, par exemple en termes d'âge ou de pouvoir.}

        %% Citations
        \subsubsection{Part des hommes cités}
            
            \paragraph{}{Le deuxième indicateur mesuré dans les articles est celui de la part des hommes cités dans le texte par rapport à l'ensemble des personnes citées, c'est-à-dire les personnes dont les propos sont rapportés par le ou la journaliste.}
            
            \paragraph{}{La citation est un indicateur de choix pour rendre compte des inégalités de représentation en termes de genre dans la presse: sur la distribution sociale des sources citées, nous pouvons évoquer entre autres l'étude de Karen Ross \cite{ross_silent_2011} qui observe, dans son article {\it "Silent Witness: News Sources, the Local Press and the disappeared Woman"}, que les femmes sont trois fois plus susceptibles d'être citées en tant que membres du public que les hommes –eux étant davantage cités en tant que représentants ou experts, de par leur profession ou leur position d'autorité. Aurélie Olivesi \cite{olivesi_les_2012} observe également cette tendance de discrimination genrée des sources citées dans les médias français, dans son étude des personnes "non expertes" citées dans {\it Le Monde}. A une échelle plus générale, les études telles que le GMMP évoqué plus haut et celle de Ross, démontrent l'intérêt de quantifier la distribution genrée des sources citées dans le discours de la presse: ce discours établit les voix reconnues comme légitimes à commenter les sujets publics et sociaux. Une mesure continue et globale de ce phénomène telle que celle que nous proposons permet donc de rendre compte le caractère hautement systématique de ce déséquilibre d'accès à la parole des femmes par rapport aux hommes.}
            
             La part des hommes cités pour un article est définie par l'équation \ref{eq:part_H_c}. Pour un document (article) $d$, la $\textrm{\it Part H cités}_d$ est simplement le nombre d'hommes cités $\textrm{\it H cités}$ divisé par le nombre total de femmes et d'hommes cités.  $\textrm{\it Part H cités}_d=0$ correspond à un article dont 100\% des citations sont attribuées à des femmes. Comme pour l'indice précédent, la moyenne journalière par source est calculée par l'équation~\ref{eq:part_H_c_source}. Cette proportion peut également être calculée par rubrique. 
            
            \begin{equation}\label{eq:part_H_c}
            \textrm{\it Part H cités}_d = \left(\frac{\textrm{\it H cités}}{\textrm{\it H cités + F citées}}\right) \in [0,1]
            \end{equation}

            \begin{equation}\label{eq:part_H_c_source}
            \textrm{\it Part H cités}(D) =\left( \frac{1}{|D|} \sum_d \textrm{\it H cités}_d \right) \in [0,1]
            \end{equation}
            
            \paragraph{}{Cet indicateur est calculé à partir du genre des personnes dont les propos sont rapportés par le ou la journaliste. Ces citations peuvent être incluses entre guillemets (par exemple: "Je pars", dit Joanne Doe), ou rapportées de manière indirecte (par exemple: Joanne Doe a dit qu'elle partait). Par exemple, dans la phrase "Jeanne D et Georges E sont membres du conseil. Jeanne D dit: 'Je souhaite démissionner'", la part des hommes cités sera de zéro, puisque seule Jeanne D sera comptée comme personne citée, et identifiée comme femme. %Cet indicateur est donc calculé selon la formule $Part\_H\_cités = \frac{H\_cités}{H\_cités + F\_citées} \in [0,1]$, où 0 correspond à un article dont 100\% des citations sont attribuées à des femmes. 
            Une explication plus détaillée de la manière dont cette mesure est mise en oeuvre se trouve en section \ref{sec:citations}.}

%%%%%%%% Données
\section{Données}
\label{sec:collecte}

    \subsection{Méthode de collecte des données}
    \label{sec:scraping}
    
        \paragraph{}{Les données utilisées dans le cadre de ce projet sont accessibles en ligne sur les sites web des journaux sélectionnés. Afin d'identifier chaque jour les nouveaux articles publiés sur ces sites, nous utilisons les comptes Twitter principaux de ces médias dont l'activité principale consiste à annoncer ces parutions et à orienter l'audience vers le site par le biais d'un lien hypertexte \footnote{Il s'agit des comptes suivants : \at LaCroix, \at Le\_Figaro, \at lemondefr, \at le\_Parisien, \at lequipe, \at LesEchos, \at libe}. Nous collectons chaque nuit grâce à des techniques de {\it scraping} l'ensemble des tweets postés la veille sur Twitter par les journaux du corpus et les URL vers lesquelles ils orientent les lecteurs et lectrices sur ce réseau social. Le nombre d'articles identifiés par ce biais pour chaque source est renseigné dans le tableau \ref{tab:infos_db}. Cette méthode ne nous paraît pas susceptible d'introduire un biais dans la sélection des articles inclus dans notre analyse.}

    \subsection{Base de données extraite}
    \label{sec:bdd}
    
      \paragraph{}{A partir des liens URL obtenus, un algorithme de {\it scraping} permet de recueillir une série d'informations pour chaque article. Chaque article parcouru donne lieu à une entrée dans notre base de données dont les champs sont les suivants: l'URL de l'article, la date de parution, le nombre de mots accessibles, les auteurs et autrices (si l'information est disponible), la source, la rubrique, le titre, le lien URL vers l'image (si image il y a) et le type d'accès (payant ou gratuit). Nous ne conservons pas le texte de l'article dans notre base de données: il est traité à la volée, et uniquement utilisé pour calculer les mesures dont nous détaillons le processus plus bas.}
   
     \paragraph{}{Afin de pouvoir procéder à des comparaisons des mesures selon les rubriques sur plusieurs sources, un dernier traitement est effectué concernant la catégorie de l'article. Chaque journal possède en effet ses propres dénominations de rubriques. Afin de les homogénéiser et de permettre la comparaison entre sources et l'agrégation des données, nous avons déterminé une liste de catégories thématiques générales à partir de l'observation des données collectées (par exemple: "Science et environnement", "Culture", "Politique", …). A chaque catégorie de la liste sont associées pour chaque journal les rubriques correspondantes qui lui sont propres. Par exemple, la rubrique "Culture" de {\it lemonde.fr} est associée avec notre catégorie générale CULTURE, et sa rubrique "Sciences" avec SCIENCE\_ET\_ENVIRONNEMENT. Les rubriques "Cinéma" et "Musique" de {\it lefigaro.fr} sont regroupées dans CULTURE. Les rubriques plus rares sont regroupées dans une catégorie "INDEFINI".}
    
%%%%%%%% Mesures
\section{Mesures}
\label{sec:methodo}

    \subsection{Mesure du taux de masculinité des mentions}
    \label{sec:mentions}
    
        \subsubsection{Méthodologie}
        \label{sec:methodo_masc_mention}
    
            \paragraph{}{Pour calculer le niveau de masculinité des personnes mentionnées dans les articles nous utilisons une méthode fondée sur le repérage des prénoms présents dans le texte des articles. Nous avons constitué, à partir des données de l'INSEE\footnote{La base de données de départ est le fichier Prénoms 2017 qui contient 32 703 prénoms d'enfants nés en France depuis 1900}, un fichier de prénoms non ambigus pour lesquels nous connaissons le nombre d'enfants filles et garçons à qui il est attribué chaque année et donc la probabilité que son porteur soit un garçon. Ce fichier contient plus de 11 000 prénoms dont les caractéristiques principales sont qu'ils sont relativement fréquents (au moins 100 occurrences dans la base prénoms de l'INSEE), qu'ils contiennent au moins 4 caractères (afin d'éviter les cas fréquents d'acronymes identiques à des prénoms courts) et qu'ils ne sont pas ambigus, c'est-à-dire qu'ils ne désignent pas autre chose qu'une personne. Pour respecter cette dernière contrainte nous avons donc supprimé de cette base de données les prénoms qui désignent aussi des pays ('Chine'), des mois de l'année ('Avril'), des couleurs ('Blanche'), des villes ('Paris') ou d'autres choses ou qualités diverses ('Justice', 'Victoire', 'Noël').}

            \paragraph{}{Nous attribuons à chaque prénom rencontré dans le texte des articles un score de masculinité égal à la probabilité que ce prénom soit donné à un garçon depuis 1900. Nous calculons ensuite la moyenne de ces scores pour chaque article que nous appelons le taux de masculinité des mentions de l'article. On peut illustrer ce calcul à partir de l'exemple suivant : le prénom Jean-Michel a un taux de masculinité de 1 selon l'INSEE alors que Loïs a un taux de 0.69, Camille de 0.25 et Maëva de 0. Un article qui mentionnerait un Jean-Michel et deux Camille aurait donc un taux de masculinité des mentions de $\frac{1 + 0.25 + 0.25}{3}=0.5$.}
    
            \paragraph{}{Dans la très grande majorité des cas, ces prénoms désignent des personnes mentionnées par le ou la journaliste dans l'article parce qu'il ou elle les a rencontrées dans un reportage ou qu'elles sont les sujets de son article. Mais il est important de garder en tête qu'il peut aussi s'agir de personnes mentionnées par d'autres personnes dans l'article ("Angela Merkel a déclaré qu'Emmanuel Macron était d'accord avec elle") ou encore, plus rarement, de la mention de prénoms désignant d'autres choses que des individus comme des lieux ou des institutions nommé•es d'après des personnes ("Le collège Henry IV").}
            
            \paragraph{}{Le score calculé dépend du nombre d'occurrences des prénoms dans un article contrairement à celui qui pourrait l'être en assignant un genre certain aux personnes citées et en comptant ces personnes. Si un•e journaliste cite trois fois plutôt qu'une le prénom d'une source dont le score de masculinité est de 0, il ou elle contribue d'autant à faire baisser l'indice de masculinité des mentions de l'article. L'indice mesure donc davantage la masculinité perçue par la lectrice ou le lecteur du fait de l'emploi de prénoms genrés que la répartition entre personnes de sexe masculin ou féminin mentionnées dans l'article.}

        \subsubsection{Évaluation}
        
        \paragraph{}{Notre algorithmique de calcul de ce taux de masculinité des mentions a été évalué sur un échantillon de 100 articles de notre base de données. Dans chacun de ces articles ont été annotés manuellement les prénoms des personnes qui y apparaissent. Un genre ("Masculin", "Féminin", "Neutre" ou "Inconnu") a été attribué également manuellement à chacun de ces prénoms en fonction des indices présents dans le texte (pronoms, accords, etc). A partir de ces informations, le taux de masculinité des mentions de référence a été calculé manuellement pour chacun de ces articles. Un prénom masculin équivalait à 1, un prénom féminin à 0, et un prénom neutre à 0.5. Ce taux de masculinité calculé manuellement a ensuite été comparé au taux de masculinité calculé par l'algorithme. Pour évaluer notre algorithme avec cet échantillon limité, nous avons effectué le test non paramétrique de Wilcoxon de rangs signés. Nous obtenons une valeur {\it p} d'environ 0.2026. L'écart-type calculé est d'environ 0.1725. Notre résultat étant supérieur au risque $\alpha$ de 0.05, nous pouvons rejeter l'hypothèse nulle selon laquelle les deux distributions (le taux calculé manuellement et le taux calculé par l'algorithme) sont significativement différents.}
    
    %%% Citations
    \subsection{Mesure de la répartition des citations selon le genre}
    \label{sec:citations}
    
        \subsubsection{Méthodologie}
        
            \paragraph{}{Le calcul de la part des hommes cités repose actuellement sur un algorithme d'extraction de citations à base de règles. Cet algorithme a été initialement développé par le {\it Discourse Lab} de l'Université Simon Fraser (Vancouver, Canada). Nous l'avons adapté et amélioré en ajoutant des règles d'extraction et en utilisant des modèles plus performants. De chaque texte est extrait un nombre de schémas linguistiques qui correspondent à l'introduction de citations. Ces schémas ont été déterminés préalablement à partir de la littérature linguistique et d'une liste de verbes de parole. A chaque citation extraite est ensuite attribué$\cdot$e son auteur ou autrice grâce à des règles soit syntaxiques (par exemple, le sujet grammatical du verbe de parole introduisant la citation), soit de proximité (le locuteur ou la locutrice cité$\cdot$e précédemment).}
            
            \paragraph{}{Les auteurs et autrices des citations d'un texte peuvent prendre plusieurs formes: il peut s'agir de noms propres ("Doanna Joe"), de groupes nominaux ("la directrice de l'agence"), de pronoms ("elle") ou de combinaisons de ces catégories ("Doanna Joe, la championne du monde"). Afin d'obtenir notre indicateur final de la part d'hommes cités dans un texte, il est nécessaire de déterminer le genre de chaque auteur et autrice d'une citation de manière automatique. Celui-ci est déterminé à partir d'un ou plusieurs des indices suivants :}
            
            \begin{itemize}
                \item la présence d'un titre genré (par exemple, "Madame" ou "M.")
                \item la présence d'un nom de métier, renseigné par le biais d'un dictionnaire de noms féminins et masculins de métiers et de fonctions, constitué manuellement à partir du guide {\it Femme, j'écris ton nom … Guide d'aide à la féminisation des noms de métiers, titres, grades et fonctions} \cite{becquer1999femme}
                \item la reconnaissance du genre du pronom ("il, "elle", \dots)
                \item  la reconnaissance d'entités nommées : d'abord, le système identifie automatiquement la présence éventuelle d'un nom de personne dans chaque auteur ou autrice de citations identifié$\cdot$e. Par exemple, le système reconnaîtra ``Doanna Joe'' comme une personne si l'auteur·ice est ``Doanna Joe, championne du monde''). Le modèle de NER ({\it Named Entity Recognition}) utilisé est celui de la librairie {\it Stanza} \cite{qi2020stanza}. De ce nom de personne est extrait le prénom, puis le genre correspondant à ce prénom à l'aide de deux dictionnaires qui attribuent un genre à chaque prénom (un dictionnaire de la librairie python {\it gender-guesser} et un autre construit manuellement à partir de la base INSEE des prénoms cité section~\ref{sec:methodo_masc_mention}).
            \end{itemize}

        \subsubsection{Évaluation}
        
            \paragraph{}{Cette méthode d'extraction de citations a été évaluée sur un corpus annoté manuellement de 54 articles de presse en français fourni par le {\it Discourse Lab}. L'évaluation obtenue est une précision d'environ 0.85, et un rappel d'environ 0.5. La tolérance est calculée sur la base du nombre de {\it tokens} en commun entre la citation détectée par l'algorithme et l'annotation manuelle. Du fait du travail d'adaptation et d'amélioration de l'algorithme, le seuil de tolérance pour cette évaluation est de 30\% de {\it tokens} en commun. Une précision de 0.85 signifie que 85\% des citation détectées sont effectivement des citations, tandis qu'un taux de rappel de 0.5 signifie que 50\% des citations contenues dans les articles ne sont pas trouvées. En ce qui concerne le genre, il peut être déterminé dans 50\% des cas de citations extraites. Les 50\% restants correspondent souvent à des pronoms impersonnels, des noms d'organisations, ou encore à des inconnues (prénom non connu, métier non genré, erreur d'extraction, \dots).}
            
            \paragraph{}{Ces résultats sont en partie dus à la méthode actuelle par règles, qui ne permet pas de détecter les cas ne correspondant pas aux schémas syntaxiques prédéfinis. Un travail en cours consiste à développer des modèles d'extraction de citations à partir de modèles plus performants à base de réseaux de neurones.}

%%% Présentation de la visualisation & Analyse sur échantillon
\section{Visualisation et observations}
\label{sec:obsv}

    \subsection{Description des jauges de {\it GenderedNews}}
    \label{sec:jauges}
        
\begin{figure}[!htb]
        \centering
        \begin{subfigure}{0.4\textwidth}
            \includegraphics[width=\textwidth]{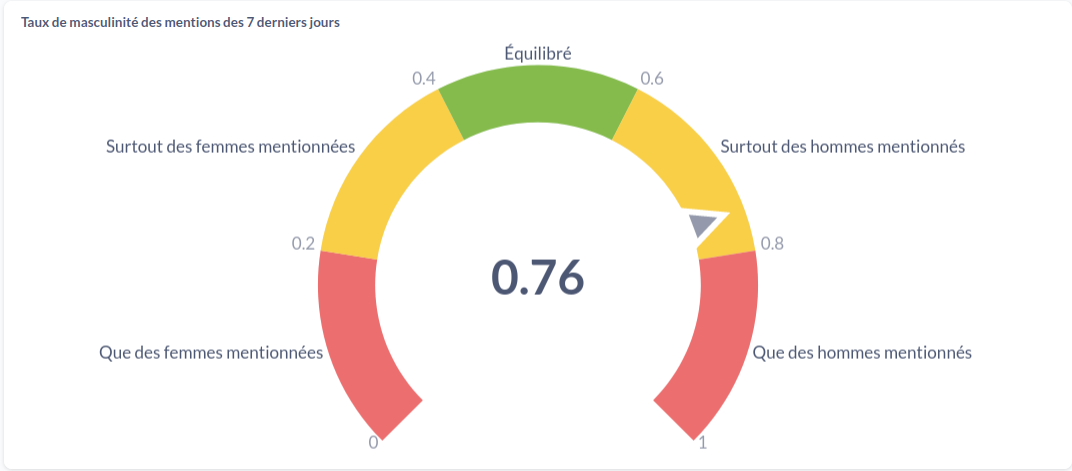}
            \caption{Taux de masculinité des mentions entre le 16/12/21 et le 23/12/21}
            \label{fig:mentionsmascsem}
          \end{subfigure}
          \hfill
          \begin{subfigure}{0.4\textwidth}
            \includegraphics[width=\textwidth]{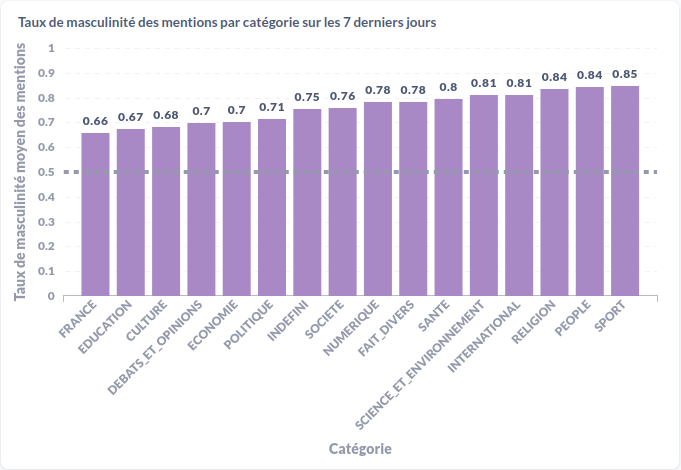}
            \caption{Taux de masculinité des mentions par catégorie entre le 16/12/21 et 23/12/21}
            \label{fig:mentionsmasccat}
          \end{subfigure}
          \hfill
          \begin{subfigure}{0.75\textwidth}
            \includegraphics[width=\textwidth]{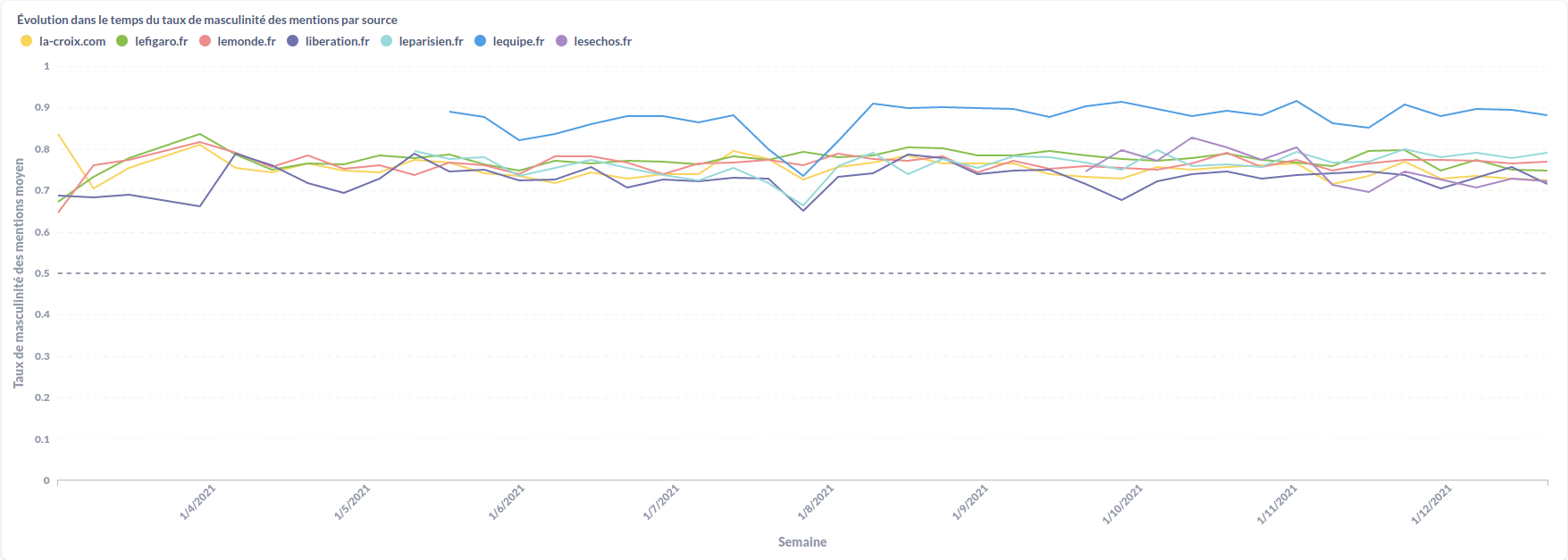}
            \caption{Taux de masculinité des mentions par source depuis le 01/03/21}
            \label{fig:mentionsmascsource}
          \end{subfigure}
        \caption{Visualisations du taux de masculinité des mentions}
        \label{fig:jaugesmentions}  
        \end{figure}

\begin{figure}[!htb]
        \centering
        \begin{subfigure}{0.4\textwidth}
            \includegraphics[width=\textwidth]{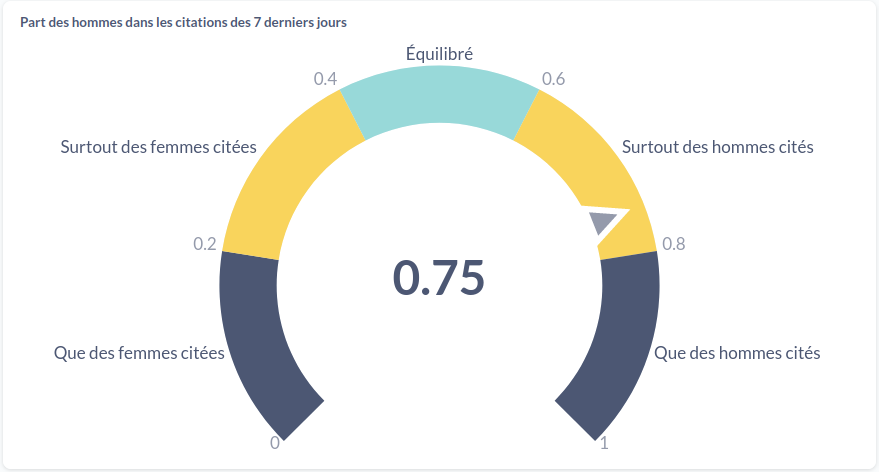}
            \caption{Part des hommes cités entre le 16/12/21 et le 23/12/21}
            \label{fig:citationssmascsem}
          \end{subfigure}
          \hfill
          \begin{subfigure}{0.4\textwidth}
            \includegraphics[width=\textwidth]{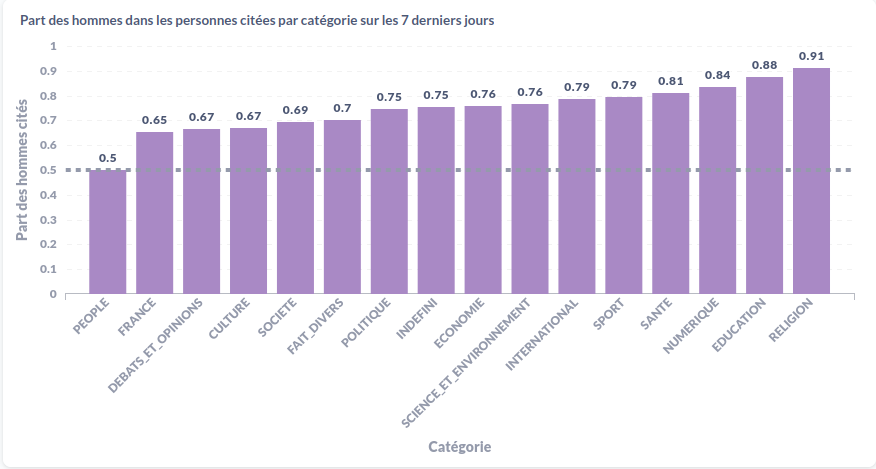}
            \caption{Part des hommes cités par catégorie entre le 16/12/21 et le 23/12/21}
            \label{fig:citationsmasccat}
          \end{subfigure}
          \hfill
          \begin{subfigure}{0.75\textwidth}
            \includegraphics[width=\textwidth]{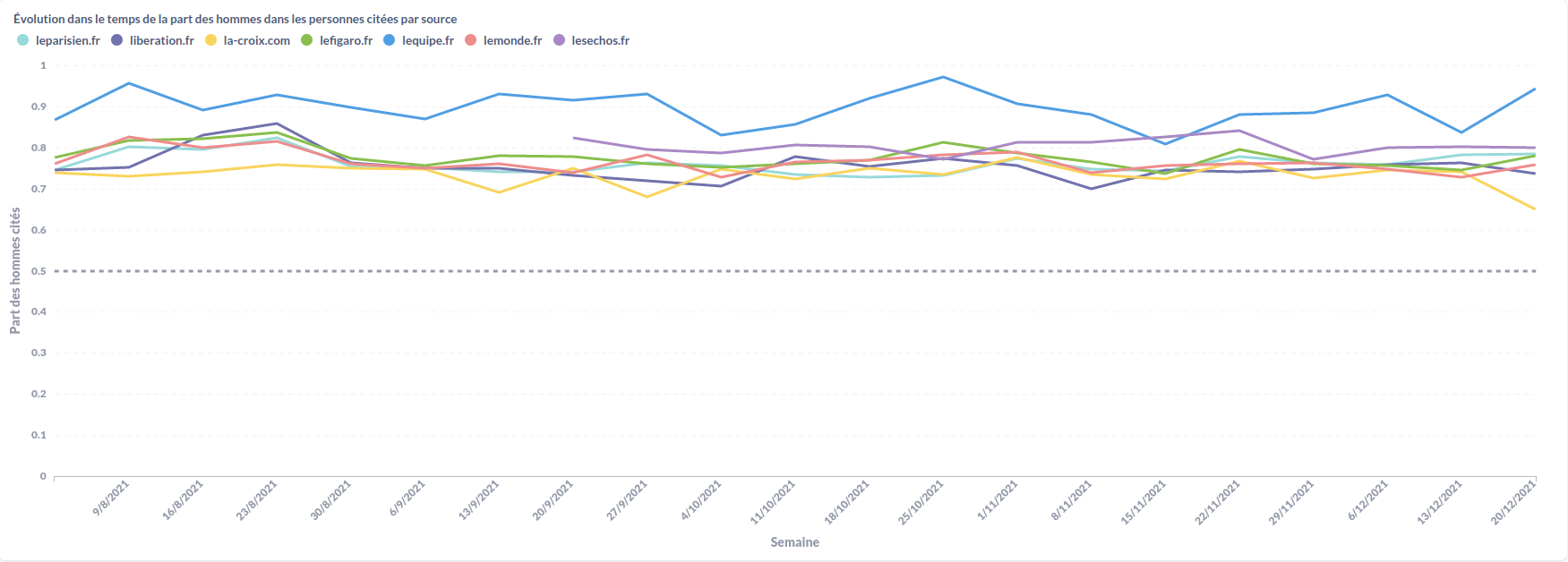}
            \caption{Part des hommes cités par source depuis le 02/08/21}
            \label{fig:citationsmascsource}
          \end{subfigure}
        \caption{Visualisations de la part des hommes cités}
        \label{fig:jaugescitations}  
        \end{figure}

        \paragraph{}{Les deux indicateurs, le taux de masculinité des mentions et la part des hommes cités, sont représentés sous trois formes différentes. Une première visualisation rend compte du niveau moyen des indicateurs sur les sept derniers jours écoulés: les figures \ref{fig:mentionsmascsem} et \ref{fig:citationssmascsem} indiquent par exemple les taux moyens dans les mentions et les citations pour la semaine du 16 décembre 2021 au 23 décembre 2021. Sont également comparés les indicateurs selon la catégorie des articles, également sur la semaine glissante. Les figures \ref{fig:mentionsmasccat} et \ref{fig:citationsmasccat} comparent respectivement le taux de masculinité des mentions et la part des hommes cités selon les catégories identifiées par l'algorithme pour la semaine du 16 décembre 2021 au 23 décembre 2021. Enfin, l'évolution globale des deux indicateurs depuis le début du projet en mars 2021 sur l'ensemble des données est représentée par les figures \ref{fig:mentionsmascsource} et \ref{fig:citationsmascsource}, mises à jour chaque lundi avec les taux moyens par source de la semaine écoulée (du lundi au dimanche).}
        
        %%%%%% Jauges pour mentions masc
        %%%%%% Jauges pour citations masc

    \subsection{Analyse d'un extrait de notre base de données}
    \label{sec:analyses}
    
    \subsubsection{Description de l'échantillon de la base de données}
    
    \paragraph{}{Cette partie sera consacrée à une analyse plus détaillée d'un échantillon de notre base de données collectée. Cet échantillon correspond à la période de deux mois du 11/07/21 au 11/09/21. Durant cette période, nous avons traité 29 556 articles. Le tableau \ref{tab:infos_sample} détaille plusieurs informations sur cet échantillon de la base de données.}
    
    \begin{table}[h]
        \centering
        \begin{tabular}{|K{2cm}|K{2cm}|K{3cm}|K{3cm}|K{4cm}|}
            \hline
            Source  & Nombre d'articles collectés & Nombre moyen de mots par article & Taux de masculinité des mentions moyen & Proportion moyenne des hommes dans les personnes citées \\
            \hline
            la-croix.com    &   2435  &   645.61  &   0.77    &   0.74 \\
            lefigaro.fr &   6199  &   421.42  &   0.79    &   0.79 \\
            lemonde.fr  &   4661  &   570.03  &   0.77    &   0.78 \\
            leparisien.fr   &   8364  &   313.45  &   0.75    &   0.77 \\
            lequipe.fr   &  6132  &   80.16   &   0.87    &   0.89 \\
            liberation.fr   &   1765   &   626.44  &   0.73    &   0.77 \\
            \hline
        \end{tabular}
        \caption{Informations sur les articles collectés entre le 11/07/21 et le 11/09/21}
    \label{tab:infos_sample}
    \end{table}
    
    \subsubsection{Résultats}
    
        \paragraph{}{L'algorithme a attribué un taux de masculinité des mentions à 25 185 de ces 29 556 articles traités, et a extrait un nombre de citations pour 20 833 articles (cela est dû au fait que l'extraction de citations a été mise en place à partir du 01/08/21).}

    %%%%%% Jauges globales échantillon
        \begin{figure}[!h]
        \centering
        \begin{subfigure}{1.0\textwidth}
            \includegraphics[width=\textwidth]{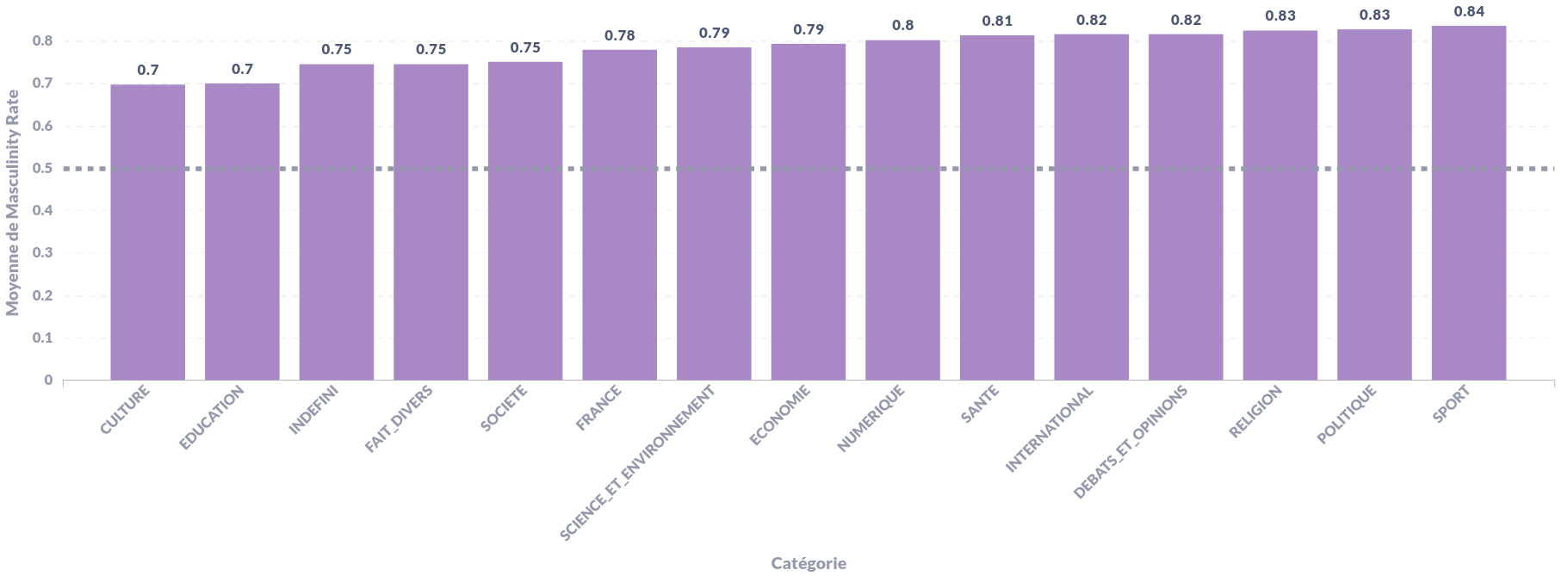}
            \caption{Moyenne du taux de masculinité des mentions par catégorie entre le 11/07/21 et le 11/09/21}
            \label{fig:mentionsmasc_sample}
          \end{subfigure}
          \hfill
          \begin{subfigure}{1.0\textwidth}
            \includegraphics[width=\textwidth]{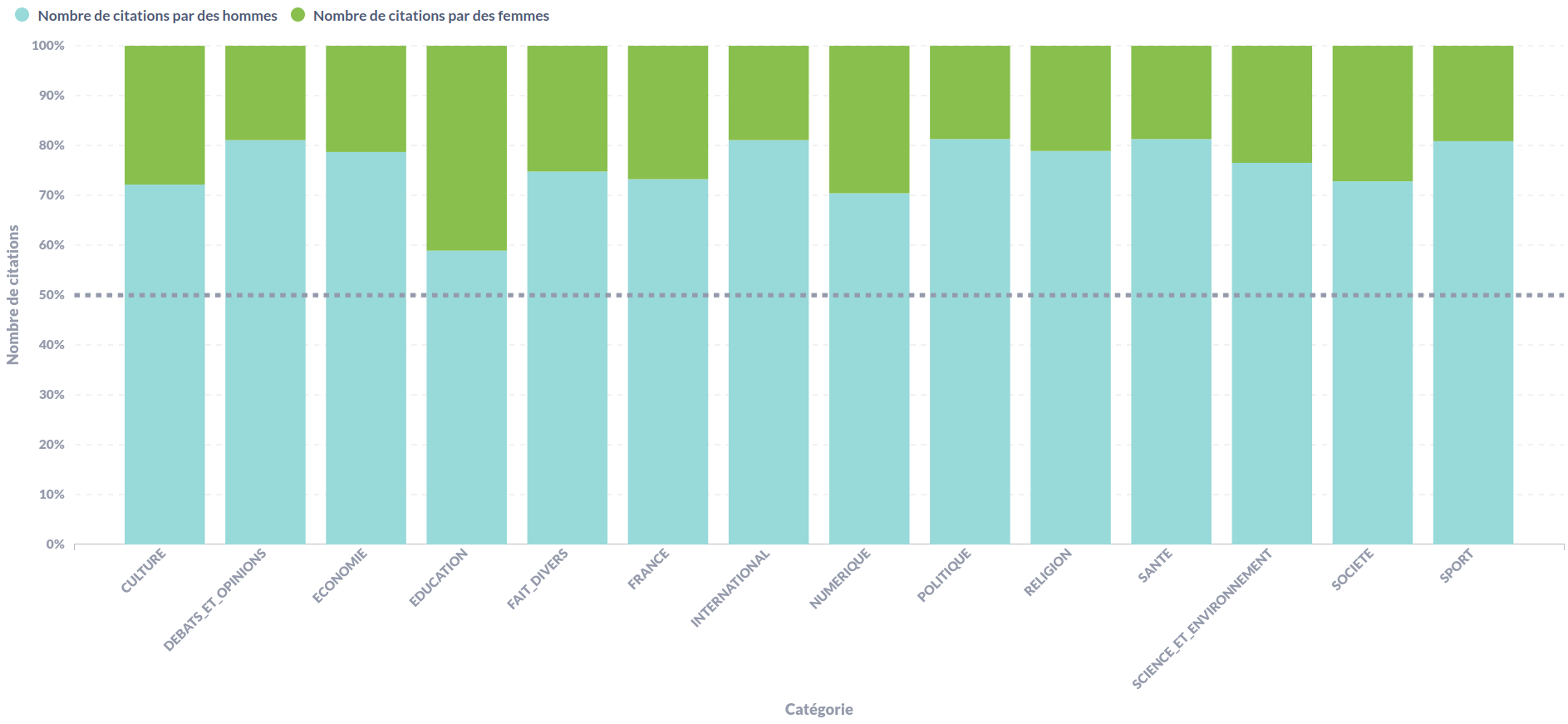}
            \caption{Proportion de citations par des hommes et des femmes par catégorie entre le 27/07/21 et 11/09/21}
            \label{fig:quotesmasc_sample}
          \end{subfigure}
          \hfill
        \caption{Indicateurs par catégorie pour la période du 11/07/21 au 11/09/21}
        \label{fig:jauges_sample}  
        \end{figure}

        \paragraph{}{Les différences entre catégories sur la période, détaillées dans les figures \ref{fig:mentionsmasc_sample} pour les mentions et \ref{fig:quotesmasc_sample} pour les citations, sont semblables à celles observées de manière hebdomadaire durant le temps du projet. De manière globale, les tendances observées placent les catégories ``SPORT'', ``POLITIQUE'', ``INTERNATIONAL'' comme celles mentionnant et citant le plus d'hommes. \`A l'inverse, les catégories ``PEOPLE'' et ``CULTURE'' citent et mentionnent le plus de femmes (les hommes étant cependant toujours majoritaires dans les mentions et les citations).}
        
        \begin{figure}[!h]
        \centering
        \includegraphics[width=0.7\linewidth]{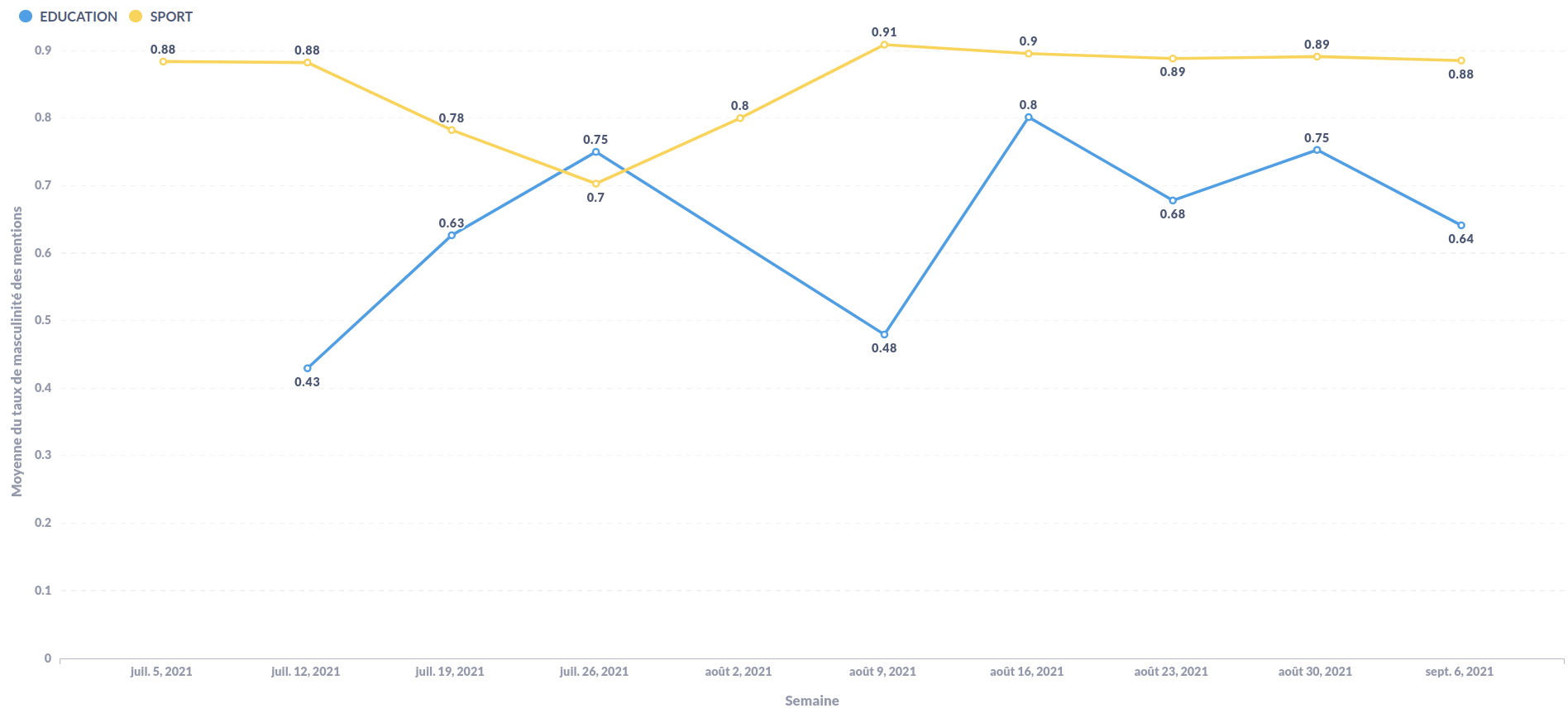}
        \caption{Moyenne du taux de masculinité des mentions par jour pour les catégories Sport et Éducation entre le 11/07/21 et le 11/09/21}
        \label{fig:mentionsSportEduc_sample}
    \end{figure}
        
        \paragraph{}{La période de l'échantillon est cependant marquée par deux phénomènes particuliers qu'il est intéressant de souligner. La figure \ref{fig:mentionsSportEduc_sample} présente le détail des moyennes de l'indicateur du taux de masculinité des mentions pour les catégories ``SPORT'' et ``ÉDUCATION''. La catégorie ``SPORT'', catégorie la plus marquée par des forts taux de présence masculine de manière générale, connaît une baisse de son taux de masculinité des mentions d'environ dix points pour les semaines entre le 19 juillet 2021 et le 8 août 2021. Cette baisse inhabituelle de la catégorie reflète la baisse globale du taux selon les titres de presse sur les mêmes semaines, observable plus haut sur la figure \ref{fig:mentionsmascsource}. Sur cette période, seule la catégorie "SPORT" connaît cette baisse importante, les autres catégories gardant un taux moyen stable par rapport au reste des données. Ce phénomène peut s'expliquer par le déroulement des Jeux Olympiques à ce moment de l'année: l'évènement a causé un afflux plus important d'articles de la catégorie "SPORT", mais a surtout provoqué une couverture médiatique plus forte du sport féminin et des médailles gagnées lors des épreuves olympiques par des femmes. La moitié des médailles françaises durant ces Jeux Olympiques ont, en effet, été remportées par des femmes.}
        
        \paragraph{}{La catégorie "ÉDUCATION", quant à elle, présente une variation importante du taux selon les semaines, ce qui est un comportement représentatif de cette catégorie depuis le début du projet. Le taux de masculinité des mentions pour la catégorie "ÉDUCATION" fluctue entre 0.43 et 0.8. On observe par ailleurs par la moyenne de ce taux sur toute la période été 2021 que la catégorie "ÉDUCATION" est celle où les femmes sont presque autant citées que les hommes (figure \ref{fig:quotesmasc_sample}). L'observation des données extraites afin de comprendre ce phénomène révèle un comportement intéressant de la catégorie. Deux angles journalistiques semblent primer quand il s'agit du sujet de l'éducation: les articles ayant un taux de masculinité élevé traitent plutôt des décisions gouvernementales concernant l'enseignement et les écoles, tandis que les articles plus paritaires traitent davantage des conditions de l'enseignement et de son public, des élèves, familles, enseignantes et enseignants.}

\section{Conclusion: limites et perspectives}
\label{sec:ccl}

    \paragraph{}{Nous avons présenté dans cet article l'approche qui est celle du site {\it GenderedNews}, par lequel nous proposons des mesures automatisées des inégalités de représentation entre hommes et femmes dans la presse écrite en ligne française. Nos résultats sont exposés dans un tableau de bord en ligne qui renseigne de manière continue la quantification de cette représentation à l'aide de deux observations principales: celle des personnes mentionnées et celle des personnes citées chaque semaine. Les résultats que nous obtenons sont cohérents avec ceux trouvés dans la littérature et les études comme le GMMP. Cette quantification concerne à ce stade du projet uniquement la présence ou non de femmes ou d'hommes, et ne constitue pas une analyse plus qualitative de cette représentation. Dans la section \ref{sec:analyses}, nous avons également proposé une exploitation plus poussée de nos données afin d'observer des comportements et événements particuliers, notamment selon les rubriques thématiques. Cette analyse nous a permis de replacer nos indicateurs dans leur contexte médiatique, ce que leur simple visualisation ne permet pas de faire.}
    
    \paragraph{}{Les méthodes de calcul des mesures de {\it GenderedNews} ont pour vocation à être améliorées et précisées. Les mesures d'évaluation décrites dans la section \ref{sec:methodo} laissent une marge de progrès. Ces résultats peuvent être améliorés grâce à des méthodes plus poussées d'apprentissage automatique. Ces méthodes nécessitent cependant un grand nombre de données. Un travail est donc en cours pour améliorer l'extraction de citations et l'extraction d'entités nommées grâce à la réalisation d'un corpus de données pour l'apprentissage. Ce système permettra également l'ajout de nouvelles mesures qui porteront notamment sur l'analyse des stéréotypes d'écriture en rapport avec le genre. Nous envisageons également d'étendre cette mesure à d'autres journaux de la presse écrite, par exemple la presse locale, la presse spécialisée et les magazines, ainsi qu'aux contenus textuels des autres médias disposant d'un site web régulièrement alimenté, comme les chaînes de télévision et les radios, et enfin aux sites web d'information dits {\it pure players}. Avec le projet {\it GenderedNews}, notre objectif est en effet de rendre compte au mieux de la systématicité et la continuité des inégalités de représentation de genre dans le plus grand nombre de sources composant le paysage des médias français.}

\section{Données et code}
\label{sec:code}
    Le code implémentant le site {\it GenderedNews} (\url{https://gendered-news.imag.fr/}) et les mesures décrites dans cet article est disponible sous licence GNU Affero General Public License v3.0 sur le dépôt Git \url{https://github.com/getalp/genderednews}.

\section*{Remerciements}
Nous souhaitons remercier le groupe d'étudiantes et étudiants du Master Informatique de Polytech Grenoble (Nhat Quang Ho, Mica Murphy, Gloria Nguena et Antoine Saget) qui ont travaillé à la construction du tableau de bord en ligne de {\it GenderedNews} et à la mise en place de l'architecture du système. Nous remercions également le {\it Discourse Lab} de l'Université Simon Fraser à Vancouver pour le partage de leur code de leur prototype de système d'extraction de citations en français. Ce travail est financé par un projet Initiatives de recherche à Grenoble Alpes (IRGA) de l'université Grenoble Alpes et a été en partie soutenu par MIAI@Grenoble-Alpes (ANR-19-P3IA-0003).

%Bibliography
\bibliographystyle{apalike}  
\bibliography{references}  
%\printbibliography

\end{document}